\pdfoutput=1

\documentclass[11pt]{article}

\usepackage{acl}

\usepackage{times}
\usepackage{latexsym}
\usepackage{hyperref}
\usepackage{multirow}
\usepackage{colortbl}
\usepackage{amsmath}
\usepackage{graphicx}
\usepackage{longtable}
\usepackage{caption}
\usepackage{subcaption}
\usepackage{float}
\usepackage{tabularx}
\usepackage{xcolor}
\usepackage{enumitem}
\usepackage[T1]{fontenc}

\usepackage[utf8]{inputenc}

\usepackage{microtype}

\title{UMass-BioNLP at MEDIQA-M3G 2024: DermPrompt - A Systematic Exploration of Prompt Engineering with GPT-4V for Dermatological Diagnosis}

\author{Parth Vashisht\thanks{* Equal Contribution}, 
  Abhilasha Lodha\footnotemark[1],
  Mukta Maddipatla\footnotemark[1], \\
  \bf{Zonghai Yao}, 
  \bf{Avijit Mitra}, 
  \bf{Zhichao Yang}, 
  \bf{Junda Wang}, 
  \bf{Sunjae Kwon},
  \bf{Hong Yu} \\ \\
  \bf{CICS, University of Massachusetts, Amherst, MA, USA} \\
  \texttt\ \{\href{mailto:pvashisht@umass.edu}{pvashisht}, \href{mailto:alodha@umass.edu}{alodha}, \href{mailto:mmaddipatla@umass.edu}{mmaddipatla}, \href{mailto:zonghaiyao@umass.edu}{zonghaiyao}\}@{umass.edu}
  \\
}

\begin{document}
\maketitle
\begin{abstract}
 
This paper presents our team's participation in the MEDIQA-ClinicalNLP 2024 shared task B. We present a novel approach to diagnosing clinical dermatology cases by integrating large multimodal models, specifically leveraging the capabilities of GPT-4V under a retriever and a re-ranker framework. Our investigation reveals that GPT-4V, when used as a retrieval agent, can accurately retrieve the correct skin condition 85\% of the time using dermatological images and brief patient histories. Additionally, we empirically show that Naive Chain-of-Thought (CoT) works well for retrieval while Medical Guidelines Grounded CoT is required for accurate dermatological diagnosis. Further, we introduce a Multi-Agent Conversation (MAC) framework and show it’s superior performance and potential over the best CoT strategy. The experiments suggest that using naive CoT for retrieval and multi-agent conversation for critique-based diagnosis, GPT-4V can lead to an early and accurate diagnosis of dermatological conditions. The implications of this work extend to improving diagnostic workflows, supporting dermatological education, and enhancing patient care by providing a scalable, accessible, and accurate diagnostic tool.~\footnote{The code is released at \href{https://github.com/parth166/M3G-ClinicalDermatology}{Github}}

\end{abstract}

\section{Introduction}
Diagnosing skin conditions demands a complex blend of visual inspection, patient history examination, and deep clinical acumen, a skill set that dermatologists spend extensive years acquiring~\cite{mangion2023medical}.
Despite the critical nature of these skills, many regions worldwide face a stark scarcity of dermatological expertise~\cite{benner2009expertise}.
Even in areas with adequate services, the demand for such specialized knowledge frequently surpasses its availability. 
The recent global health crisis has also expedited the shift towards remote clinical diagnostics and treatments, further highlighting the challenges in diagnosing skin diseases~\cite{behar2020remote}.
These challenges include the scarcity of dermatological expertise and the need to accommodate asynchronous patient interactions, including e-visits, emails, and messaging platforms, to ensure continuity and quality of care.

In response to these challenges, recent advancements in Artificial Intelligence (AI), particularly through the development of large language models (LLMs), offer promising solutions to significantly support dermatologists by enhancing clinical diagnosis and treatment processes~\cite{mcduff2023towards, singhal2023towards, tu2024towards}.
Moreover, AI facilitates asynchronous patient services, offering a cost-effective and convenient alternative to traditional methods.
Previous works have primarily utilized deep learning for tasks such as skin lesion classification ~\cite{Udritoiu2020SkinDC,Esteva2017DermatologistlevelCO,Brinker2019ACN}, and dermatopathology~\cite{HEKLER201979,10.1111/bjd.18026} focusing predominantly on dermoscopic images ~\cite{10.1007/978-3-642-40763-5_50}. 
These efforts, however, have relied on image-only models, indicating a need for broader applications.

Our research aims to extend the capabilities of AI in dermatology by diagnosing skin diseases and devising appropriate treatment plans based on patients' dermatological images, queries, and medical histories. 
This approach mirrors the diagnostic process of dermatologists, who rely on high-quality images and comprehensive patient histories to make informed decisions. 
Although previous studies have explored fine-tuning models on multimodal data (e.g., SkinGPT \cite{zhou2023skingpt4}, and MedBLIP \cite{chen2023medblip}, our task is particularly challenging due to data availability and image quality limitations, reflecting real-world constraints where high-quality data is either scarce or expensive to obtain.

Studies by OpenAI \cite{nori2023capabilities} and Microsoft \cite{nori2023generalist} have demonstrated that generalist foundation models, such as GPT-4, can surpass specifically fine-tuned medical models on various medical benchmarks by employing specialized prompting strategies. 
Building on these insights, our research leverages both the textual and visual capabilities of GPT-4, targeting the specific task of dermatology diagnosis and treatment.

The diagnostic process for skin lesions or conditions requires meticulous evaluation and is informed by methodologies such as dermatoscopy \cite{panagoulias2024dermacen}, which enables dermatologists to observe skin abnormalities in greater detail. 
Dermatologists usually follow a common guideline for assessing skin lesions, emphasizing the importance of visual descriptors like shape, size, color, texture, and pattern in differential diagnosis. 
Inspired by these practices, we have integrated advanced Chain-of-Thought (CoT) techniques with visual features to create medical guidelines tailored for GPT-4V, enhancing its diagnostic precision.
This enables the model to emulate dermatologists' diagnostic process.

Furthermore, our research integrated a Multi-Agent Conversation (MAC) Framework \cite{tao2024chainofdiscussion,wu2023autogen,unknown}, which involves multiple AI agents that generate additional context and critiques for various candidate skin conditions. These agents collaborate, debate, and consolidate their findings to determine the most accurate skin disease diagnosis from the candidates identified from the retrieval step. 
This, therefore, introduces a level of dynamic interaction and comprehensive analysis that mirrors the complex decision-making process in clinical dermatology. \\

Hence, our contributions are twofold:
\begin{itemize}[itemsep=1pt, parsep=1pt]
    \item We deploy GPT-4V within a novel retrieval and re-ranking framework, critically evaluating the effectiveness of various prompting strategies. These include both naïve prompts and those meticulously crafted based on detailed medical guidelines (CoT), across different stages of our setup. This exploration aims to highlight the adaptability and precision of GPT-4V in simulating the diagnostic reasoning of dermatologists.
    \item We explore the Multi-Agent Conversation (MAC) Framework in the context of clinical dermatology, examining its potential to enrich the diagnostic process. Through this discussion, we identify and delineate the framework’s strengths and limitations, offering insights into its applicability and performance in accurately diagnosing skin diseases.
\end{itemize}

\begin{figure*}
  \includegraphics[width=\textwidth]{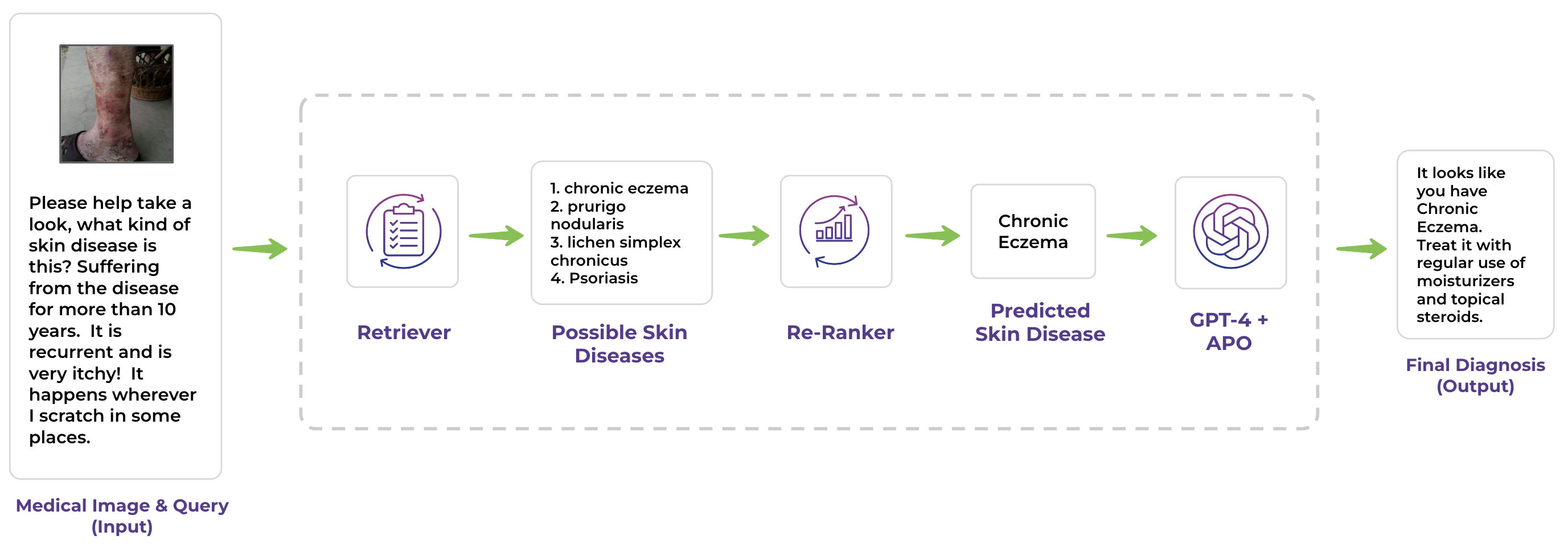}
  \caption{Overview of the AI-assisted dermatology diagnosis pipeline, from initial patient input through to the GPT-4V generated final diagnosis and treatment plan.}
  \label{methodology}
\end{figure*}

\section{Related Work}
The interdisciplinary fusion of Artificial Intelligence (AI) and dermatology has spawned a myriad of approaches to enhance the diagnosis of skin conditions. Historically, these approaches have often treated diagnosis as a classification task, with literature extensively documenting the use of convolutional neural networks (CNNs) and other deep learning architectures like ResNets for lesion classification from dermoscopic images, which are typically limited to dermatological clinics due to image acquisition constraints \cite{BA2022156}, \cite{9250855}.

Recent advancements have moved beyond traditional clinic-bound methods, exploring the utility of clinical images for broader classifications \cite{Groh2024}, such as skin cancer and onychomycosis \cite{9706441}. While these efforts have made significant contributions to disease diagnosis, they have not fully addressed the generative and comprehensive nature of clinical diagnosis, which encompasses treatment planning and patient interaction beyond mere classification. \cite{Jeong2022-ql}

The evolution of large language models (LLMs) has significantly widened the scope of AI applications in healthcare. LLMs like PubMedBERT \cite{pubmedbert} and BioGPT \cite{10.1093/bib/bbac409} have been fine-tuned on extensive corpora of medical literature, achieving state-of-the-art performance in tasks ranging from biomedical reasoning to question-answering. In the realm of domain-adapted LLMs, models like Meditron and Med-PALM have demonstrated remarkable capabilities in language understanding and generation, setting new benchmarks across biomedical datasets \cite{chen2023meditron70b,singhal2022large}.

With the advent of multimodal models, integrating visual and textual data has further refined AI applications in medical domains. \cite{info:doi/10.2196/55508} Vision-language models such as Med-CLIP, Med-BLIP, and Llava-Med have exhibited promising results in image-text retrieval, zero-shot classification, and even multimodal conversations, respectively \cite{chen2023medblip,li2023llavamed}. Specifically, in the context of dermatology, the Skin-GPT4 model \cite{zhou2023skingpt4} represents a pioneering effort in creating a multimodal setup tailored for skin disease identification and patient interaction.

Studies have shown that generalist foundation models like GPT-4 \cite{Pillai2024.01.24.24301743}, with their expansive knowledge bases and specialized prompting techniques \cite{chen2023enhancing}, outperform domain-specific models such as Med-PALM on various medical benchmarks \cite{nori2023capabilities}. GPT-4's application in dermatology, particularly in melanoma identification and medical exam question answering, underscores its potential as an assistive tool for educational and diagnostic purposes \cite{10.1093/ajcp/aqae030}, \cite{mishra2024synfacedit}, \cite{Yang2023.10.26.23297629}.

Our research builds upon these foundations, employing GPT-4's multimodal capabilities (GPT-4V) to enhance dermatological diagnostic processes. By integrating Chain-of-Thought (CoT) techniques and a Multi-Agent Conversation (MAC) Framework \cite{tao2024chainofdiscussion,wu2023autogen,unknown}, we aim to emulate the complex decision-making process of dermatologists, enriching the GPT-4V's ability to generate diagnostic and treatment plans from multimodal data. This work not only taps into the multimodal analytical strength of GPT-4V but also seeks to optimize the model's performance in a domain where the nuances of patient history and visual inspection are paramount.

Thus, our contribution to the field involves the innovative use of GPT-4V within a retrieval and re-ranking framework, leveraging both naïve and medically informed CoT prompting strategies.

\section{Methodology}
Our methodology delineates the comprehensive approach we adopted to address the task of multimodal medical answer generation. This process involves two primary stages: retrieving potential diagnoses and the ranking of these to identify the most probable skin condition and treatment plan.

Task Description:
The objective of our research is to develop a system capable of diagnosing a possible skin condition and recommending a corresponding treatment plan based on a patient's medical query and associated image. To accomplish this, we propose a two-step pipeline consisting of a retrieval module followed by a ranker module. Specifically:
\begin{itemize}
\item Retrieval Module: This component extracts a list of possible skin conditions from the given image and medical query.
\item Ranker Module: This module's task is to select the most accurate skin condition diagnosis from the list generated by the retrieval module.
\end{itemize}

\noindent Our overall methodology is mentioned in Figure \ref{methodology}.

\subsection{Retrieval Module}
\label{subsec:retreival}
The Retrieval Module is the initial phase of our diagnostic approach, reflecting the dual aspects a dermatologist considers when evaluating a medical condition: visual inspection and patient history. 
Inspired by recent works such as MedGENIE~\cite{frisoni2024generate}, along with others~\cite{yu2022generate,zhang2023merging, su2022context}, we leverage LLMs as strong context generators instead of traditional retrieval methods, such as keyword-based methods (e.g., BM25~\citet{robertson2009probabilistic}), vector-similarity-based methods (e.g., ColBERT~\citet{khattab2020colbert}), and some internet tools (e.g., Google API). 
In this retrieval step, we treat LLMs as a knowledge base~\cite{singhal2023large} to generate potentially valuable information for subsequent steps.
Our module employs two distinct strategies:

\subsubsection{Context-Independent Retrieval (Image-Only)}
Recognizing scenarios where comprehensive medical context (patient's medical history and medical queries) may not be readily available, we engage in context-independent retrieval. This approach leverages GPT-4V to identify possible skin conditions based solely on image data.
We compared this model's performance against a widely-used online AI tool, First Derm~\footnote{https://firstderm.com/}

\subsubsection{Context-Dependent Retrieval (Image + Context)}
The inclusion of medical context is pivotal for accurate diagnosis. 
Particularly, incorporating details about systemic conditions and patient history can significantly influence differential diagnosis, a critical aspect of clinical dermatology.
To this end, we utilize CoT prompting, a technique that simplifies complex problems into manageable objectives, enabling the model to address the larger task. \\

\noindent Within the Context-Dependent Retrieval, we experiment with two strategies:
\begin{itemize}[itemsep=1pt, parsep=1pt]
    \item Naive CoT: Here, GPT-4V is instructed to methodically analyze all relevant information from the images and medical query before generating a list of potential skin conditions. This process mimics the step-by-step procedural thinking a dermatologist might employ.
    \item Expert Guidelines Grounded CoT: This approach involves crafting prompts based on the  \href{https://www.ncbi.nlm.nih.gov/books/NBK378786/}{Clinical Guidelines} that dermatologists follow, encapsulating a generic framework for skin disease diagnosis. Such frameworks typically comprise patient history, visual inspection, and differential diagnosis. Our Expert-CoT strategy emphasizes key visual characteristics like the lesion's shape, color, size, location, and texture. By integrating this data, the module produces a detailed list of differential diagnoses needed to enhance the model's diagnostic precision further.~\footnote{The specific prompts utilized for both the Naive CoT and the Expert Guidelines Grounded CoT strategies are detailed in Table \ref{tab:cot_retrieval}}

\end{itemize}

\subsection{Re-Ranker Module}
After the retrieval module identifies potential skin conditions, the re-ranker module is critical in our diagnostic pipeline. Its primary objective is to meticulously refine the preliminary list, pinpointing the diagnoses with the highest probability of accuracy. 
To achieve this, we experiment with four re-ranking strategies:
\begin{enumerate}[itemsep=1pt, parsep=1pt]
    \item Naive Chain of Thought (CoT)
    \item Expert Guidelines Grounded CoT with Context
    \item Expert Guidelines Grounded CoT without Context
    \item Multi-Agent Conversation Framework
\end{enumerate}

These structured approaches enable a systematic evaluation of the candidate's conditions, ensuring that the decision-making process mirrors the analytical and deductive reasoning of a dermatologist.
The specific prompts utilized for these three CoT techniques are presented in Tables \ref{tab:naive_ranking}, \ref{tab:with_context_ranking}, and \ref{tab:without_context_ranking}.
\noindent The MAC framework is explained with an example in the Appendix \ref{sec:appendix}.

\subsubsection{Naive Chain of Thought (CoT)}
In the Naive CoT approach, GPT-4V is initially instructed to analyze the patient's medical query and the associated images to extract relevant information. Subsequently, each candidate skin condition retrieved from section \ref{subsec:retreival} is assigned a score ranging from 1 to 10, where 1 signifies the least probable and 10 denotes the most probable condition. The model identifies the most probable disease based on the scores and analysis. The prompt is mentioned in Table \ref{tab:naive_ranking}.

\subsubsection{Expert Guidelines Grounded CoT with Context}
This method employs a sophisticated strategy by utilizing prompts meticulously designed around the \href{https://www.ncbi.nlm.nih.gov/books/NBK378786/}{Clinical Guidelines} followed by dermatologists. These guidelines encapsulate a comprehensive visual assessment of the affected area, scrutinizing distinct characteristics such as shape, size, color, location, and texture (listed in Table \ref{tab:with_context_ranking}). Within this framework, GPT-4V is initially directed to conduct an analysis of the patient's condition, incorporating insights drawn from their medical query. Subsequently, the patient's images undergo visual examination using the defined guidelines, using which relevant features are extracted. The final step involves considering a list of possible skin conditions (retrieved from section \ref{subsec:retreival}) and systematically ruling them out based on the gathered insights and visual inspections to identify the most probable skin condition from the set of candidates.

\subsubsection{Expert Guidelines Grounded CoT without Context}
This approach omits the user query, focusing exclusively on the visual examination of dermatological conditions as per established \href{https://www.ncbi.nlm.nih.gov/books/NBK378786/}{guidelines}. Utilizing GPT-4V, an initial step involves the generation of a detailed visual description, drawing upon ten specified visual features essential for dermatological assessment (as outlined in Table \ref{tab:without_context_ranking}). Subsequently, each candidate's skin condition is described visually, emphasizing distinguishing features aligned with the visual guidelines. A comparative analysis is then conducted between the visual descriptions of the candidates and the initial image description, and a score ranging from 1 to 10 is assigned based on the level of match (1 being the lowest match and 10 the highest). The most probable candidate, determined by the highest score in the comparative analysis, is selected as the diagnosis.

\begin{figure*}[ht]
  \centering
  \includegraphics[width=1.0\linewidth, height=1.0\textheight, keepaspectratio]{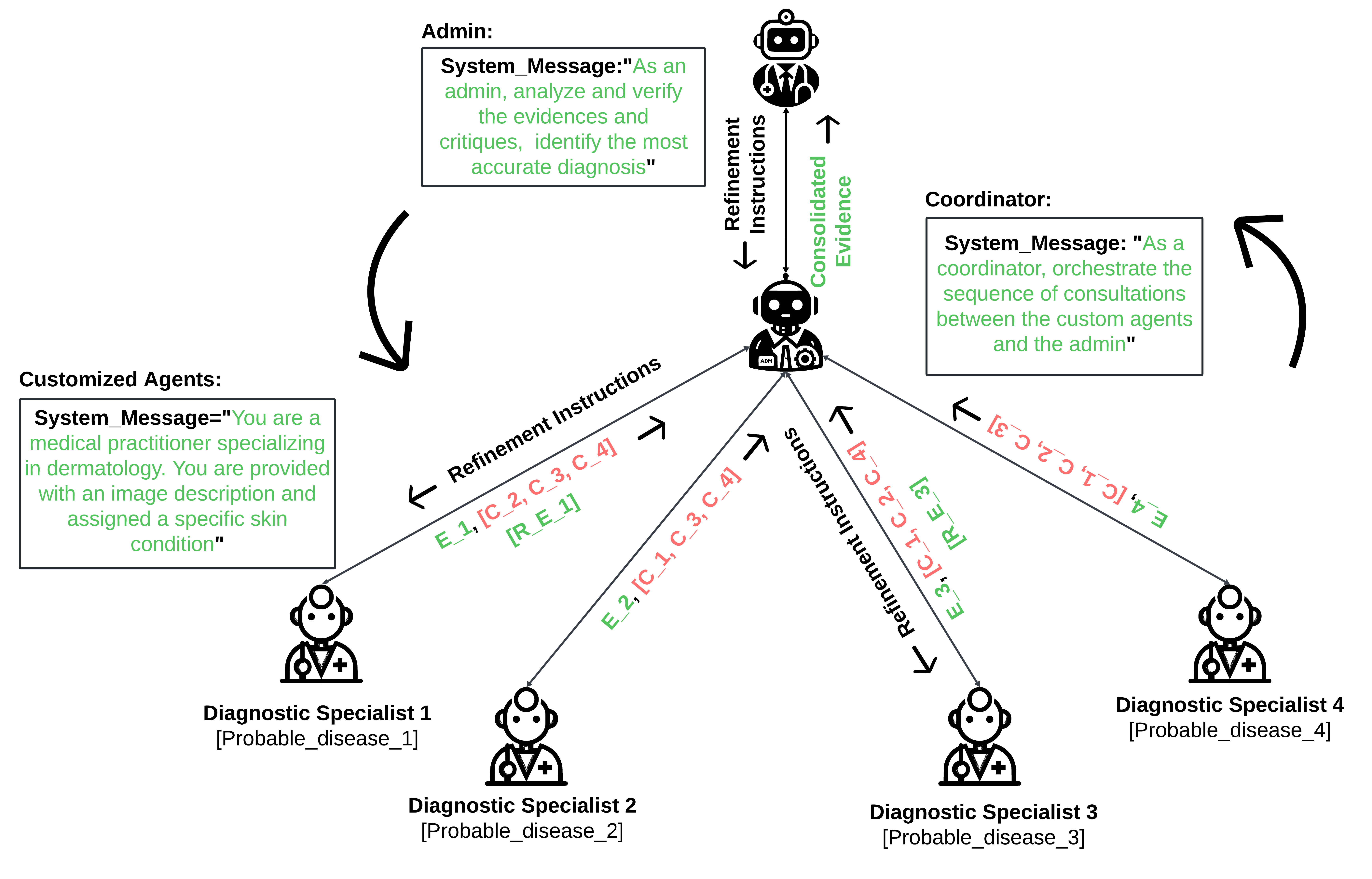}
  \caption{Multi-Agent Conversation (MAC) Setup }
  \textit{\small Note: E\textsubscript{1} corresponds to the evidence supporting probable\_disease\_1 generated by Diagnostic\_Specialist\_1. C\textsubscript{2}, C\textsubscript{3}, and C\textsubscript{4} are the critiques for probable\_disease\_2, 3 \& 4 generated by Diagnostic\_specialist\_1. Based on the Re-definement Instructions from the Admin, the diagnostic specialist returns Redefined evidence(R\_E\_1 )}
  \label{fig:mac}
\end{figure*}

\subsubsection{Multi-Agent Coversation Setup}
Inspired by the recent various applications of the Multi-Agent Conversation framework in the general and medical domains~\cite{wu2023autogen,tao2024chain}, we also implement a multi-agent conversation framework for our re-ranker module (Figure~\ref{fig:mac}).
This framework involves multiple AI agents, each specializing in a different aspect of dermatology diagnosis. These agents collaborate, debate, and consolidate their findings to identify the most accurate diagnosis, mirroring the collaborative approach often seen in medical panels. This multi-agent setup not only enriches the model’s diagnostic capabilities but also introduces a level of dynamic interaction and comprehensive analysis that mirrors the complex decision-making process in clinical dermatology. Moreover, acknowledging performance gain and consistency improvement obtained using critique-based refinement in large language models, we incorporate feedback generation as an objective of multi-agent debate followed by refinement. \\
\noindent The main components of our multi-agent setup are defined in Table \ref{tab:mac_roles}.

\noindent \textbf{Process Flow}
\vspace{-4pt}
\begin{itemize}[itemsep=1pt, parsep=1pt]
    \item Assignment and Analysis
    \begin{itemize}
        \item The Coordinator assigns a distinct probable disease to each Diagnostic Specialist based on the case study and list of probable diseases.
        \item Each specialist analyzes the case study, provides evidence supporting their assigned disease and critiques the applicability of other diseases.
    \end{itemize}
    \item Compilation and Presentation of Findings
    \begin{itemize}
        \item After receiving inputs from all specialists, the Coordinator compiles the evidence and critiques.
        \item The compiled information is presented to the Admin for evaluation. 
    \end{itemize}
    \item Evaluation and Revision
    \begin{itemize}
        \item The Admin reviews the evidence and critiques, identifying areas where additional clarity or strengthening is needed.
        \item If necessary, the Admin requests revisions from specific specialists to enhance their evidence based on critiques.
    \end{itemize}
    \item Final Diagnosis
    \begin{itemize}
        \item With the revised evidence, the Admin conducts a final review to determine the most accurate diagnosis.
        \item The process concludes once the Admin confirms the diagnosis.
    \end{itemize}
\end{itemize}

\begin{table*}
\begin{tabular}{|p{0.20\textwidth}|p{0.70\textwidth}|}
\hline
\textbf{Roles} & \textbf{Tasks} \\ \hline
\textit{Coordinator} & Orchestrate sequence of consultations. Assign diseases to specialists and manage communications. \\ \hline
\textit{Admin} & Evaluate evidence and critiques for accuracy. Request evidence enhancements and finalize the diagnosis. \\ \hline
\textit{Agent} & Analyze case study, advocate for one disease. Provide evidence and critique alternative diagnoses. \\ \hline
\end{tabular}
\caption{Tasks for each role in a multi-agent setup}
\label{tab:mac_roles}
\end{table*}

\subsection{Aligner}
The Aligner Module represents the final step in our diagnostic process, focusing on optimizing the model's output to ensure it aligns with clinical standards and expectations. This involves adjusting the prompt to refine the model's language and structure, aiming to emulate the concise, informative style characteristic of professional medical advice. The optimization process is guided by analyzing real doctor responses in the dataset, identifying key elements such as terminology, format, and the inclusion of essential diagnostic and treatment information. The goal is to produce a diagnosis and treatment plan that not only accurately identifies the patient's condition but also provides actionable, understandable advice. This module highlights our commitment to bridging the gap between AI-generated content and the practical needs of clinical practice, ensuring that the output is not only technically accurate but also clinically relevant and usable in real-world medical contexts.

Recent work has introduced aligners to assist LLMs in generating harmless outputs~\cite{ji2024aligner}, a concept previously applied in the style transfer domain to map model outputs to desired forms (e.g., formality style transfer~\cite{rao2018dear,yao2021improving}). 
When using third-party APIs like GPT, where updating the model's weights is not an option, recent methods have explored the use of Automatic Prompt Optimization (APO) to improve prompts, assuming access to training data and an LLM API~\cite{prasad2022grips,pryzant2023automatic}. 
Recent studies have also applied APO in the clinical domain to assist doctors in generating better note-generation prompts~\cite{yao2023physicians}. 
Inspired by these efforts, we use human responses from training data as APO's training input, allowing the LLM to derive appropriate aligner prompts to facilitate the final step of style transfer.
The final prompt generated by APO can be found in Table~\ref{tab:apo}.

\subsection{Evaluation}
The evaluation of our pipeline is dependent on the accuracy metric. Accuracy is defined individually for each component.
\begin{equation}
\text{Accuracy} = \frac{\text{Number\ of\ retrieved\ GT}}{\text{Total\ number\ of\ data\ points}}
\end{equation}
\\
\textit{Number of retrieved GT}: Total number of examples for which ground truth skin condition was present in the retrieved list of candidate skin conditions. \\
\textit{Total number of data points}: Total number of examples for which ground truth skin condition is known. We skip all those examples for which ground truth is not known. In the validation we have 47 examples for which ground truth is known and a total of 56 examples.

Acknowledging the fact that a same skin condition can have multiple names, we implemented GPT-Eval as an evaluator to identify if two skin conditions are similar or not. Our evaluation strategy employs a rule-based approach to assess the similarity between two skin conditions, "A" and "B", according to four predefined rules (refer Table \ref{tab:evaluation_rules}). These rules incorporate name identity, synonymity, common root condition, and shared effects and causes to determine similarity systematically. This method addresses the complex nature of dermatological conditions by providing a structured framework that considers linguistic, clinical, and etiological aspects of skin diseases.

\section{Results}
\subsection{Retrieval Module} 
\begin{table}[htbp]
\centering
\begin{tabular}{|l|l|c|}
\hline
\multicolumn{1}{|c|}{\textbf{Retrieval Strategies}} & \multicolumn{1}{c|}{\textbf{Methods}} & \textbf{Accuracy} \\ \hline
\multirow{2}{*}{\textit{Context Independent}}  & First Derm  & 0.468085 \\ \cline{2-3}
                                      & GPT-based   & 0.595744 \\ \hline
\multirow{2}{*}{\textit{Context Dependent}}    & Naive CoT   & 0.851063\\ \cline{2-3}
                                      & Expert CoT  & 0.744680 \\ \hline
\end{tabular}
\caption{Comparison of Retrieval Strategies and their Accuracy}
\label{tab:retrieval-strategies}
\end{table}

\noindent The accuracy scores, as reported in Table \ref{tab:retrieval-strategies}, reveal significant insights into the efficacy of each strategy employed within our Retrieval Module. \\
Firstly, we observed that Context-Independent Retrieval, which relies exclusively on image data, resulted in lower accuracy when compared to Context-Dependent strategies. This indicates that the absence of medical context limits GPT-4V's ability to identify potential skin diseases accurately. \\
Conversely, Context-Dependent Retrieval exhibited superior results. By incorporating medical queries along with images, this method provides a richer context to GPT-4V, leading to more precise retrieval of potential skin conditions. It appears that the additional contextual data plays a pivotal role in enhancing the model's diagnostic capabilities. \\
When comparing the two strategies within the Context-Dependent Retrieval, Naive CoT outperformed Expert CoT. This may initially seem counterintuitive, given that Expert CoT is grounded in medical guidelines, which one would expect to yield better results. However, our analysis suggests that the Naive CoT strategy's ability to generate a broader range of potential candidates contributed to its higher accuracy. In contrast, the Expert CoT strategy, which employs differential diagnosis principles, likely eliminated some candidates during the retrieval phase, potentially leading to decreased accuracy. \\
From these observations, we hypothesize that differential diagnosis, while not as effective in the initial retrieval phase, may be better suited to the re-ranking phase of our diagnostic pipeline. The re-ranking phase requires a systematic evaluation to differentiate between closely related skin conditions, aligning with the differential diagnosis's intrinsic nature. Therefore, the nuanced approach of systematically eliminating similar conditions could prove beneficial in the subsequent stage, where precision is paramount.

\subsection{Re-Ranker Module}

\begin{table}[ht]
\begin{tabular}{|l|c|c|}
\hline
\textbf{Methods}                                                             & \multicolumn{1}{l|}{\textbf{\begin{tabular}[c]{@{}l@{}}Top-2\\ Accuracy\end{tabular}}} & \multicolumn{1}{l|}{\textbf{\begin{tabular}[c]{@{}l@{}}Top-1\\ Accuracy\end{tabular}}} \\ \hline
Naive CoT                                                                   & 0.553191                                                                                      & 0.425531                                                                                      \\ \hline
\begin{tabular}[c]{@{}l@{}}Medical Guidelines\\ (image+context)\end{tabular} & 0.617021                                                                                      & 0.531915                                                                                      \\ \hline
\begin{tabular}[c]{@{}l@{}}Medical Guidelines\\ (image only)\end{tabular}    & 0.553191                                                                                      & 0.446808                                                                                      \\ \hline
\end{tabular}
\caption{Comparison of Re-Ranker Strategies and their Accuracy}
\label{tab:ranker_strategies}
\end{table}

The re-ranker module is important in refining the initial list of potential diagnoses obtained from the retrieval module. The metrics used to evaluate the performance of our re-ranker module are:
\begin{itemize}
    \item \textbf{Top-2 Accuracy:} This metric reflects the model's ability to include the correct diagnosis within its top two predictions from the candidate conditions identified in the retrieval phase.
    \item \textbf{Top-1 Accuracy:} This is the precision with which the model identifies the correct diagnosis as its first and final choice from all possible conditions.
\end{itemize}

As illustrated in Table \ref{tab:ranker_strategies}, the evaluation of our re-ranker strategies reveals several insights. \\
The Naive CoT and Medical Guidelines (image only) strategies exhibit comparable performance, with both Top-2 and Top-1 accuracies closely aligned. This suggests that even without the medical context, the model can leverage visual cues to a degree of effectiveness. \\
A notable increase in accuracy is observed with the use of Medical Guidelines alongside context (image + patient's query). Incorporating the patient's medical history and associated query, in conjunction with differential diagnosis techniques as outlined in Table \ref{tab:with_context_ranking}, enhances the model's discriminatory power. This aligns with our hypothesis that the systematic approach of differential diagnosis—filtering through similar skin conditions—proves more efficacious in the re-ranking phase.

\subsection{MAC}
\begin{table}[ht]
\begin{tabular}{|l|c|}
\hline
\textbf{Methods}    & \textbf{Accuracy} \\ \hline
\textbf{MG-GR} & 0.53333                \\ \hline
\textbf{MAC}  & 0.73333               \\ \hline
\end{tabular}
\caption{Re-Ranking - top 1 Accuracy using MAC. Here MG-GR is the (Medical Guidelines Grounded Re-Ranker.}
\label{tab:mac_scores}
\end{table}

\noindent The multi-agent conversation setup significantly outperforms the traditional top-1 re-ranking strategy, exhibiting a substantial improvement of nearly 20 percentage points. This enhancement was observed across 15 distinct examples where the number of potential solutions retrieved varied between three and five. We propose that the key mechanism driving this enhanced accuracy is the system’s critique-based conversational framework. Within this framework, each participating agent is subject to a rigorous process of critique and feedback from other agents. This collaborative interaction encourages continuous reassessment and refinement of each agent’s initial diagnoses and the evidence they present. Consequently, this iterative process likely contributes to more precise and reliable diagnostic outcomes, as each agent integrates insights gained from the critiques to adjust and improve their reasoning and conclusions.

\subsection{Aligner Module}
\begin{table}[ht]
\begin{tabular}{|l|c|}
\hline
                    & \textbf{DeltaBleu} \\ \hline
\textbf{Before APO} & 0.944723                \\ \hline
\textbf{After APO}  & 2.737657               \\ \hline
\end{tabular}
\caption{DeltaBleu scores before and after Automatic Prompt Optimization (APO)}
\label{tab:apo_results}
\end{table}

An important evaluation metric for the competition is the deltableu score. The "DeltaBLEU" score is a variation of the BLEU (Bilingual Evaluation Understudy) score, which is a widely used metric for evaluating the quality of text that has been machine-translated from one language to another. The BLEU score measures the correspondence between a machine's output and that of a human, providing a quantitative assessment of translation accuracy. We leverage Automatic Prompt Optimization (APO) to learn a set of rules that bootstraps our prediction and align the responses. Table \ref{tab:apo_results} shows the bleu score improvement by leveraging the rules learned by APO. The learned rules are mentioned in the Table \ref{tab:apo}.
\begin{itemize}[itemsep=1pt, parsep=1pt]
    \item Before Alignment: "Based on the visual descriptions, it seems like the most probable condition is Chronic Eczema. I recommend applying topical steroids and moisturizers regularly for treatment."
    \item After Alignment: "Consider Eczema, which should manifest similarly on both sides. Treat it with regular use of moisturizers and topical steroids."
    \item Ground Truth: "Should be happening on both sides.  Think of Eczema."
\end{itemize}

\section{Discussion and Conclusion}
In our study, we systematically explored the merits of various prompting strategies within an information retrieval-based dermatology diagnostic framework. By evaluating these strategies through the lens of accuracy metrics, we found that a naive Chain of Thought (CoT) strategy effectively simulates a retrieval module typical of information retrieval systems. This approach is adept at returning a sufficient number of candidate diagnoses, setting a foundational stage for further analysis. Our findings underscore the importance of including patient history and contextual information in clinical dermatology to enhance diagnostic accuracy.

For the nuanced task of re-ranking diagnostic candidates, our research indicates that a more refined CoT strategy is necessary. Specifically, prompts that incorporate expert guidelines prove critical in conducting differential diagnoses, yielding superior performance in top-1 and top-2 diagnostic outcomes.

Furthermore, we introduce the novel Multi-Agent Critique (MAC) framework, which incorporates agent-based critique and feedback, and has the potential to perform differential diagnosis and refine it's output using feedback.

\section{Limitations and Future Work}
Our current pipeline does not fully comply with stringent data protection regulations, such as the Health Insurance Portability and Accountability Act (HIPAA). Despite Azure's availability of a HIPAA-compliant hosting option, our framework has not been fully aligned with these regulatory standards. The imperative to protect patient data privacy and ensure security is paramount in clinical applications. Our findings suggest that deploying a local model might offer a more privacy-centric approach. However, achieving satisfactory performance with local deployment necessitates further research and development. This limitation underscores the critical need to balance privacy considerations with technological efficacy, especially in the sensitive context of healthcare.

An additional dimension of our study pertains to the inherent variability in the performance of the prompting strategies, attributed to the high temperature setting utilized during GPT-4's open-ended generation tasks. This element of randomness introduces inconsistencies in the model's responses. We hypothesize that employing over-sampling techniques from GPT-4, coupled with self-consistency prompting, could mitigate these inconsistencies and enhance the overall effectiveness of the diagnostic process.

Furthermore, the MAC framework's practical application presents challenges, notably in the seamless integration of inter-agent communication. Our observations point to instances where the system failed due to unexpected behaviors during these interactions, highlighting the complexities of implementing such a framework effectively. Additionally, the MAC study has been conducted on a small set of 15 samples for which the number of retrieved candidates are in the range of 3 to 5 with the MAC system failing for number of candidates greater than 5. This was because of limited context length window with GPT-4 model. Additionally, each call to the agent is financially prohibitive as the number of candidates increase since more rounds of conversations are needed. Such challenges underscore the need for further research and development to refine and optimize the MAC framework for clinical diagnostic applications.

Additionally, given the challenging nature of the dataset with unclean/missing context for a lot of examples, correct bench marking cannot be assumed, but this study can serve as a potential lower bound of GPT-4V's performance on the complex task of clinical dermatology.

\bibliography{clinical_nlp24}
\bibliographystyle{acl_natbib}

\appendix

\section{Appendix}
\label{sec:appendix}

\subsection{Multi-Agent Conversation (MAC) Case Study}

In this section, we present an exemplar case study of a debate facilitated by the Multi Agent Chat (MAC) system. The dialogues exemplified in Table~\ref{tab:MAC_1} to Table~\ref{tab:MAC_6} illustrate the dynamic interaction between diagnostic specialists and admin agents within our MAC framework. 

The ensuing discussion is prefaced with prompts which have successfully generated the anticipated outcomes, showcasing the MAC system's adeptness. We provide a series of prompts, as detailed in Table~\ref{prompt} and the expert guidelines to ensure that the communication trajectory remains aligned with the system's strategic objectives.

Through the presented case study, we aim to elucidate the capabilities of the MAC system's in the context of clinical diagnosis and the efficacy of its prompts in steering the group chat among various agents to achieve coherent, goal-oriented dialogue. 

\paragraph{Custom Agent (Diagnostic Specialist) Prompt} is  to guide the model in adopting the role of a diagnostic specialist. It aims to facilitate the identification of salient features of a skin condition based on clinical observations that align with the designated skin disease. Utilizing these features and the provided details regarding the disease progression, the model is tasked with generating evidence to substantiate the diagnosis of the specified skin disease. 

\paragraph{Coordinator Prompt} is utilized to guide the model to play the role of a coordinator, orchestrating the conversation among the agents, collecting the generated evidences and critiques, consolidating them and passing them to the Admin agent for further analysis and diagnosis. This agent is essential to ensure smooth transition in between the agents.

\paragraph{Admin Prompt} is designed to instruct the model to play the role of Admin, the admin is a head doctor, who first meticulously assesses the quality of generated evidences and critiques. Then the Admin is tasked with guiding the relevant specialists to refine their evidence in light of the critiques received, thereby enhancing the robustness of the diagnostic case. This iterative process of evaluation and refinement underscores the Admin's pivotal role in ensuring the accuracy and integrity of the final diagnosis, highlighting the significance of expert oversight in collaborative medical diagnostics.

\begin{table}[ht]%
\begin{tabularx}{0.48\textwidth}{p{7.4cm}} %
\hline
\small{\textbf{Diagnostic Specialist Prompt}} \\ 
\hline
\small{As 'Rick', you are a medical practitioner specializing in dermatology.
    You are provided with an image description and assigned a specific skin condition, your role is to tune the image description to match with the disease.
    
    Once done you can use the image description to generate a detailed report providing evidence that supports this diagnosis. Afterwards critique each of the other probable diseases by explaining why they do not fit the case study as well as your assigned diagnosis does. 
    Ensure clarity and comprehensiveness in your analysis and critiques.
    }\\
\hline

\small{\textbf{Coordinator Prompt}}\\ 
\hline
\small{As the 'Coordinator', your primary responsibility is to oversee the diagnostic process. You will receive a clinical observation of a skin disease, a case study along with a list of probable diseases. Your task is to assign each Diagnostic Specialist a unique probable disease to advocate for, based on the provided details. You'll ensure that each specialist receives all necessary information to perform their analysis effectively. Finally, gather the tuned image descriptions, evidences and critiques from the specialists and present them to the Admin for final evaluation. Your role is crucial for maintaining efficient communication and organization among the specialists.}\\
\hline

\small{\textbf{Admin Prompt}} \\
\hline
\small{As the 'Admin',your objective is to evaluate the evidence and critiques provided by the Diagnostic Specialists majorly based and aligned to the image description since solely depending on the case study can be tricky to determine the most probable disease for a given case study. Initiate your process by assessing the quality of each critique. Seek consensus among the critiques to strengthen the evidence for a particular diagnosis. You may need to instruct Diagnostic Specialists to refine their evidence based on your findings. Through a structured discussion with the Coordinator and the Diagnostic Specialists, lead the team towards agreeing on a final, most suitable diagnosis for the case study.} \\
\hline

\hline
\end{tabularx}
\caption{Prompts for Multi Agent Chat.}
\label{prompt}
\vspace{-4mm}
\end{table}

\subsection{Guidelines and Instructions for Multi-Agent Chat}
The instructions and expert guidelines are most crucial for the MAC module, since it is a comprehensive briefing of the objectives that the system aims to fulfill. This prompt delineates not only the sequence of actions requisite for task execution but also the intricacies of inter-agent transitions, thereby charting the entirety of procedural flow. Furthermore, it encompasses a set of critical guidelines mandating adherence to principles of clarity and precision, alongside the seamless exchange of information among pertinent agents. These directives are imperative to avoid miscommunication and ensure that all interactions remain aligned with the task’s end goals. The Task Prompt is furnished to the GroupChatManager, serving as the catalyst for activating dialogues among specialized agents within the framework. This structured approach is pivotal in harmonizing the collective efforts of diverse agents, thus optimizing the overall functionality and efficacy of the MAC system.

\subsection{Multi-Agent Chat - Example}
The entire chat is accessible in our GitHub repository for reference. In this section, we present selected excerpts from the Multi-Agent Chat to illustrate the flow and demonstrate the system's capabilities: \\
\\
\textbf{Set-up of the Chat:}\\
\textbf{Patient Query (with the context of disease progression)}: "The skin condition, as shown in the images, presents widespread erythematous patches with violaceous hues across the leg. The patient has multiple crusted plaques and erosions, with sizes varying from a few millimeters to several centimeters. Some lesions have a serpiginous border, suggesting an active edge. The skin's texture looks lichenified in some places, indicating chronicity, and scaling is evident across various regions, signaling some level of dryness and exfoliation. Some patches have merged, forming a larger area of affected skin. Signs of excoriations are present, most likely due to itching, and scattered pustules can also be observed."\\
\textbf{Probable Diseases}: "Prurigo nodularis, Chronic eczema, Psoriasis, Lichen simplex chronicus, Allergic or irritant contact dermatitis"\\
\textbf{Clinical Observation of the Skin Condition}: "The skin condition, as shown in the images, presents widespread erythematous patches with violaceous hues across the leg. The patient has multiple crusted plaques and erosions, with sizes varying from a few millimeters to several centimeters. Some lesions have a serpiginous border, suggesting an active edge. The skin's texture looks lichenified in some places, indicating chronicity, and scaling is evident across various regions, signaling some level of dryness and exfoliation. Some patches have merged, forming a larger area of affected skin. Signs of excoriations are present, most likely due to itching, and scattered pustules can also be observed."

\textbf{Ground Truth}: Chronic Eczema

\textbf{Excerpt 1 - GroupChat Initialization:}
The group chat is initiated by the admin. The task, meticulously crafted for our use case, is provided to the chat manager who then follows the outlined steps.

\textbf{Excerpt 2 - Evidence \& Critiques:}
After the coordinator assigns probable diseases to each diagnostic specialist, they are sequentially called to generate supportive evidence and critique other possibilities.

\textbf{Excerpt 3 - Consolidated Evidence \& Critiques:}
Once the diagnostic specialists have processed their assigned diagnoses, the coordinator gathers and consolidates the evidence and critiques for each disease.

\textbf{Excerpt 4 - Admin Refinement Instructions:}
This consolidated evidence and critiques are reviewed by the admin, who assesses them and may request further information. The admin provides instructions for specialists to refine their evidence, aiming for a more accurate diagnosis.

\textbf{Excerpt 5 - Enhanced Evidence:}
Based on the refinement instructions from the Admin, the designated agent is tasked with enhancing their evidence to better support their diagnosis.

\textbf{Excerpt 6 - Final Diagnosis:}
Based on the refined evidence, the admin determines the most relevant final diagnosis.

\begin{table*}[ht]
\centering
\begin{tabular}{|p{0.9\textwidth}|}
\hline
\textbf{Excerpt - 1:} \\
\\
**Chat Initialisation with the Admin:**\\
Admin (to chat\_manager):\\
The objective is to facilitate a systematic discussion to diagnose a patient's skin condition, based on the clinical observation of the skin condition (\textcolor{blue!50}{The skin condition, as shown in the images, presents widespread erythematous patches with violaceous hues across the leg. The patient has multiple crusted plaques and erosions, with sizes varying from a few millimeters to several centimeters. Some lesions have a serpiginous border, suggesting an active edge. The skin's texture looks lichenified in some places, indicating chronicity, and scaling is evident across various regions, signaling some level of dryness and exfoliation. Some patches have merged, forming a larger area of affected skin. Signs of excoriations are present, most likely due to itching, and scattered pustules can also be observed.}), case study (\textcolor{blue!50}{Please help take a look, what kind of skin disease is this? Suffering from the disease for more than 10 years.  It is recurrent and is very itchy!  It happens wherever I scratch in some places.}) and a list of probable diseases (\textcolor{blue!50}{['prurigo nodularis', 'chronic eczema', 'psoriasis', 'lichen simplex chronicus', 'allergic or irritant contact dermatitis']}). Give equal emphasis to the clinical observation as well as the symptoms and the disease progression to ensure accuracy. Assume that no further information or biopsy will be conducted.

"""\\
**Process Overview:** \\
"""\\
"""\\
**Detailed Steps:** \\
"""\\
"""\\
**Rules for Effective Process Management:**\\
"""\\

**Once "TERMINATE" has been returned no further conversation must happen**\\
\\
\hline
\end{tabular}

\caption{Excerpt 1 : Group chat initialisation}
\label{tab:MAC_1}
\end{table*}

\begin{table*}[ht]
\centering
\begin{tabular}{|p{0.9\textwidth}|}
\hline
\textbf{Excerpt - 2:} \\
\\
Rick (to chat\_manager):

As Diagnostic Specialist Rick,

\textcolor{blue!50}{Assigned Diagnosis: Prurigo Nodularis.}

\textcolor{teal}{Supporting Evidence for Prurigo Nodularis:} The indications of chronic scratching or rubbing like lichenification suggest that the rash could be Prurigo Nodularis. The patient's description of the condition as being very itchy and recurrent over a span of more than a decade also aligns with this diagnosis. In addition, the presence of scattered pustules can also be seen in cases of Prurigo Nodularis.

\textcolor{red}{Critiques for Other Diseases:}\\
1. \textbf{Chronic Eczema}: This condition also presents an itchy rash that can become lichenified from chronic scratching, but usually has a more defined pattern of eruption that is not described here.\\
2. \textbf{Psoriasis}: While this condition presents chronic plaques, they usually exhibit a characteristic silvery scale. No such description is provided here.\\
3. \textbf{Lichen Simplex Chronicus}: While this condition is characterized by lichenification, it generally affects a specific region rather than having a widespread distribution as described here.\\
4. \textbf{Allergic or Irritant Contact Dermatitis}: These conditions generally present rapidly after contact with an offending substance, which does not fit the chronic, decade-long presentation described here.\\
\\
\textcolor{teal}{**CALL COORDINATOR** to validate completion.} \\
\hline
\end{tabular}
\caption{Excerpt 2: Evidence \& Critiques by Diagnostic Specialist Rick.}
\label{tab:MAC_2}
\end{table*}

\begin{table*}[ht]
\centering
\begin{tabular}{|p{0.9\textwidth}|}
\hline
\textbf{Excerpt - 3:} \\
\\
Coordinator (to chat\_manager):

Thank you for your findings, Michael.

As the Coordinator, I acknowledge the completion of Michael's analysis regarding 'allergic or irritant contact dermatitis'.

\textcolor{teal}{**Compiling Findings**}

I'll now compile and categorize the generated evidences and critiques for each potential disease:\\
1. \textbf{Prurigo Nodularis}\\
- \textcolor{teal}{Supporting Evidence}: Chronic scratching or rubbing and recurrent itchiness over a decade. Presence of scattered pustules.\\
- \textcolor{red}{Consolidated Critiques}: More defined pattern of eruption for Chronic Eczema is missing. No silver scales as in Psoriasis. More localized presentation expected as in Lichen Simplex Chronicus. No acute flare-ups post exposure as in Contact Dermatitis.\\
2. \textbf{Chronic Eczema}\\
- \textcolor{teal}{Supporting Evidence}: Widespread and merging erythematous patches, lichenified skin at some regions and itchiness wherever scratched.\\
- \textcolor{red}{Consolidated Critiques}: Prurigo Nodularis typically presents hard, itchy lumps. Silvery scales typical to Psoriasis missing. Lichen Simplex Chronicus is usually localized.\\
3. \textbf{Psoriasis}\\
- \textcolor{teal}{Supporting Evidence}: Widespread plaques, lichenified skin and scaling.\\
- \textcolor{red}{Consolidated Critiques}: In Prurigo Nodularis itching is more generalized. No typical eczema characteristics like weeping, oozing. Lichen Simplex Chronicus usually is localized.\\
4. \textbf{Lichen Simplex Chronicus}\\
- \textcolor{teal}{Supporting Evidence}: Lichenification of skin due to chronic rubbing, intense itchiness, especially in localized areas, plus the serpiginous border.\\
- \textcolor{red}{Consolidated Critiques}: Larger areas of affected skin unlike Prurigo Nodularis. No oozing or crusting unlike Chronic Eczema. Silvery scale of Psoriasis missing. Contact Dermatitis usually presents acute symptoms post exposure.\\
5. \textbf{Allergic or Irritant Contact Dermatitis}\\
- \textcolor{teal}{Supporting Evidence}: Chronic dryness, exfoliation, and itching response to certain irritants.\\
- \textcolor{red}{Consolidated Critiques}: Doesn't describe nodules typical to Prurigo Nodularis. Not enough weeping and crusting for Chronic Eczema. No silvery scaly plaques like Psoriasis. Lichen Simplex Chronicus is usually localized.\\
\\
As the Coordinator, I present the compiled evidence to the Admin for a final evaluation on this patient's skin condition. \\
\hline
\end{tabular}
\caption{Excerpt 3: Consolidated Evidences \& Critiques}
\label{tab:MAC_3}
\end{table*}

\begin{table*}[ht]
\centering
\begin{tabular}{|p{0.9\textwidth}|}
\hline
\textbf{Excerpt - 4:} \\
\\
Admin (to chat\_manager):

As the Admin,

Thank you, Coordinator, for compiling the findings. 

Firstly, it's noticeable that the given evidences for "Allergic or Irritant Contact Dermatitis" and "Psoriasis" are relatively weak compared to other diagnoses. I agree with the critiques that the chronicity and absence of typical defining characteristics such as acute flare-ups in the case of dermatitis and silvery scales in the case of psoriasis make these diagnoses less likely.

Focusing on "Prurigo Nodularis", the evidence is plausible but the image description does not mention the typical nodular lumps that are characteristic of this disease.

"Lichen Simplex Chronicus" has supportive evidence that strongly aligns with chronic itching and lichenification, yet the widespread presence of the disease contradicts the usual localized occurrence of this condition.

"Chronic Eczema" also aligns well with the majority of described symptoms, including itching, chronicity, and lichenification. The absence of strong objections in the critiques and aforementioned aligning symptoms gives strength to this diagnosis.

Considering all compiled evidence and critiques, it seems that \textcolor{blue}{Chronic Eczema} might be the most likely diagnosis given the available information. However, some uncertainty remains due to overlapping symptoms with other conditions and the lack of additional diagnostic tests.

Diagnostic Specialist Sam, I would like you to enhance your evidence for 'Chronic Eczema' in light of the critiques provided by other specialists. Please refer to the following critiques and \textcolor{teal}{provide more specific details that distinguish Chronic Eczema from other conditions}:

1. Prurigo Nodularis: \textcolor{blue!50}{Your analysis could benefit from addressing whether or not the absence of nodules is indeed conclusive evidence against this diagnosis.}

2. Lichen Simplex Chronicus: \textcolor{blue!50}{Can you further explain the distinguishing factors between these two conditions? Specifically, consider the details regarding distribution and impact of itch-induced scratching.}

3. Identification or ruling out of 'Allergic or Irritant Contact Dermatitis': \textcolor{blue!50}{Please provide more info that can make this differentiation clearer.} \\
\hline
\end{tabular}
\caption{Excerpt 4: Admin Refinement Instructions}
\label{tab:MAC_4}
\end{table*}

\begin{table*}[ht]
\centering
\begin{tabular}{|p{0.9\textwidth}|}
\hline
\textbf{Excerpt - 5:} \\
\\
Sam (to chat\_manager):

As Diagnostic Specialist Sam,

\textcolor{blue!50}{Enhanced Evidence for 'Chronic Eczema'}:

To revisit and strengthen my diagnosis supporting Chronic Eczema, let's address the critiques:

1. \textcolor{blue!50}{Absence of nodules in Prurigo Nodularis}: Prurigo Nodularis is characterized by hard, itchy nodules which may be paired with lichenification because of chronic scratching. However, such nodules are not explicitly reported in this clinical presentation. Instead, we note widespread erythematous patches with various sizes and serpiginous borders, a pattern more consistent with chronic eczema.

2. \textcolor{blue!50}{Distinguishing features between Chronic Eczema and Lichen Simplex Chronicus}: Though both conditions show lichenification due to chronic scratching, they do have differing behaviors. Lichen Simplex Chronicus usually exhibits itself in one or two specific regions of the body, whereas Chronic Eczema can affect larger, more widespread areas as described in this clinical case. Thus, the widespread distribution here lends more credence to a diagnosis of Chronic Eczema rather than the typically localized Lichen Simplex Chronicus.

3. \textcolor{blue!50}{Differentiating between Chronic Eczema and Allergic or Irritant Contact Dermatitis}: Contact Dermatitis generally surfaces as an acute flare-up following exposure to a particular substance and often resolves once the irritant or allergen is avoided, whereas Chronic Eczema's cause is multifactorial - influenced not only by external irritants but also by internal factors, such as the patient's immune response. Furthermore, Chronic Eczema exhibits a distinctive pattern of flares and subsiding inflammation over time. This history of enduring for over a decade and recurrent nature of the skin condition directs more towards Chronic Eczema.\\
\\
Overall, despite sharing common symptoms like itching and skin alterations with the mentioned conditions, Chronic Eczema appears to fit best given the specifics of the condition's distribution and chronicity. \\
\hline
\end{tabular}
\caption{Excerpt 5: Enhanced Evidence}
\label{tab:MAC_5}
\end{table*}

\begin{table*}[ht!]
\centering
\begin{tabular}{|p{0.9\textwidth}|}
\hline
\textbf{Excerpt - 6:} \\
\\
Admin (to chat\_manager):

As the Admin,

Thank you, Sam, for the enhanced evidence and Coordinator for your facilitation.

Taking into account the evidences and critiques from all Diagnostic Specialists and the enhanced evidence provided by Sam, I conclude that in the absence of any further diagnostic tests or additional information, the most accurate diagnosis among the provided probable diseases for this case is \textcolor{blue}{Chronic Eczema}. The patient's long term history, reported symptoms like itching wherever the patient scratches, the recurrent nature of the condition, and clinical information like widespread erythematous patches all point towards Chronic Eczema. \\
\\
\textcolor{teal}{Please **TERMINATE** the conversation, Coordinator.}\\
\\
\hline
\end{tabular}
\caption{Excerpt 6: Final Diagnosis}
\label{tab:MAC_6}
\end{table*}

\subsection{Dermatology Guidelines}
\noindent When a dermatologist evaluates a skin condition, they typically follow a systematic approach that involves several areas.
\begin{itemize}[itemsep=1pt, parsep=1pt]
    \item Patient History: Look at the "User Query" to extract relevant context that will help in accurate diagnosis of skin conditions.
    \item Visual Inspection: The initial step involves a thorough visual examination of the affected area.
\end{itemize}
For visual inspection, the dermatologist looks at the following features and creates a list of possible skin conditions that match the visual features.
\begin{enumerate}[itemsep=1pt, parsep=1pt]
    \item Size: What is the size of the skin lesions? Is it small or large?
    \item Shape: What is the shape of the lesions?
    \item Color: What is the color of the skin lesions?
    \item Location: Where is the skin lesion or rash located?
    \item Distribution Pattern: What is the distribution pattern, is it localized or widespread?
    \item Existence of symmetry: Are the lesions symmetric?
    \item Borders: Do the edges of the lesion appear sharp, irregular, or blurred?
    \item Elevation: Is the lesion is flat, raised, or depressed below the skin surface?
    \item Texture: Does the surface quality looks smooth/scaly/rough/soft/hard.
\end{enumerate}

Pattern Recognition: Dermatologists are trained in recognizing patterns that certain skin conditions commonly present. These patterns, combined with the other collected information, help in forming a preliminary diagnosis. \\
\\
Differential Diagnoses: Based on the evaluation, the dermatologist will consider a list of possible conditions (differential diagnoses) and rule them out one by one, based on the evidence and test results. \\
\\

\begin{table*}[ht!]
\begin{tabular}{|p{0.3\textwidth}|p{0.6\textwidth}|}
\hline
\textbf{Naive CoT} & Act as an expert dermatologist. \\
& You are provided with a dermatology case with images. You are tasked to create a list of possible skin conditions for the given case. \\
& \\
& \textit{Instructions:} \\
& 1: Look at the images and the medical query and see what relevant information you can extract from the medical query that can be useful in diagnosis. \\
& 2: Create a possible list of skin conditions. \\
& \\
\hline
\textbf{Medical Guidelines based CoT} & You are provided with a dermatology case with images. \\
& For this case, you are provided with some images and additional user query. You are asked to give a diagnosis for this scenario. 
Act as a dermatologist. Refer the guidelines below and follow the guidelines to generate the diagnosis. \\
& \\
& \textit{Guidelines:} \\
& When a dermatologist evaluates a skin condition, they typically follow a systematic approach that involves several key steps. \\
& \\
& \textit{Visual Inspection:} The initial step involves a thorough visual examination of the affected area. \\
& The dermatologist looks at the: \\
& \quad 1. Size \\
& \quad 2. Shape \\
& \quad 3. Color: The color (red, brown, black, blue, white) and whether it's uniform. \\
& \quad 4. Location of the lesion or rash. \\
& \quad 5. Distribution Pattern (localized/widespread) \\
& \quad 6. Existence of symmetry (yes or no) \\
& \quad 7. Borders: The edges of the lesion—are they sharp, irregular, or blurred? \\
& \quad 8. Elevation: Whether the lesion is flat, raised, or depressed below the skin surface. \\
& \quad 9. Texture: The surface quality (smooth, scaly, rough, soft, hard). \\
& \quad 10. Pattern Recognition: Dermatologists are trained in recognizing patterns that certain skin conditions commonly present. \\
& These patterns, combined with the other collected information, help in forming a preliminary diagnosis. \\
& \quad 11. Consideration of Differential Diagnoses: Based on the evaluation, the dermatologist will consider a list of possible conditions (differential diagnoses), and rule them out one by one, based on the evidence and test results. \\
& \quad 13. Create a list of possible candidates after the above steps. \\
& \\
\hline
\end{tabular}
\centering
\caption{CoT based Retrieval}
\label{tab:cot_retrieval}
\end{table*}

\begin{table*}[ht]
\centering

\begin{tabular}{|p{0.9\textwidth}|}
\hline
Act as an expert dermatologist. \\
You are provided with a dermatology case with images and associated medical query. You are tasked to choose the most probable skin condition from the set of candidates. \\
\\
Medical Query: \\
\textit{<query>} \\
\\
Candidates: \\
\textit{<candidates>} \\
\\
Instructions: \\
1: Look at the images and the medical query and see what relevant information you can extract from the medical query that can be useful in diagnosis. \\
2: Give a score to each candidate skin condition in the range of 1 - 10 with 1 being the least probable and 10 being the most probable. \\
3: Choose a single most probable disease. If there is a tie in scores for the most probable conditions, pick a single skin condition between those candidates at random and return. \\
\hline
\end{tabular}
\caption{Naive CoT for Re-Ranker}
\label{tab:naive_ranking}
\end{table*}

\begin{table*}[ht]
\centering
\begin{tabular}{|p{0.9\textwidth}|}
\hline
Act as an expert dermatologist. You are provided with a dermatology case. For this case, you are provided with some images a user query and list of candidates. \\
\\
User Query: \textit{<query>} \\
\\
Candidates: \textit{<candidates>} \\
\\
Guidelines: \\
When a dermatologist evaluates a skin condition, they typically follow a systematic approach that involves several areas. \\
Patient History: Look at the "User Query" to extract relevant context that will help in accurate diagnosis of skin conditions. \\
Visual Inspection: The initial step involves a thorough visual examination of the affected area. \\
\\
For visual inspection, the dermatologist looks at the following features and for each, the dermatologist creates a list of possible skin conditions that show such visual features. \\
1: Size: What is the size of the skin lesions? Is it small or large? \\
2: Shape: What is the shape of the lesions? \\
3: Color: What is the color of the skin lesions? \\
4: Location: Where is the skin lesion or rash located? \\
5: Distribution Pattern: What is the distribution pattern, is it localized or widespread? \\
6: Existence of symmetry: Are the lesions symmetric? \\
7: Borders: Do the edges of the lesion appear sharp, irregular, or blurred? \\
8: Elevation: Is the lesion is flat, raised, or depressed below the skin surface? \\
9: Texture: Does the surface quality looks smooth/scaly/rough/soft/hard. \\
\\
Pattern Recognition: Dermatologists are trained in recognizing patterns that certain skin conditions commonly present. These patterns, combined with the other collected information, help in forming a preliminary diagnosis. \\
\\
Differential Diagnoses: Based on the evaluation, the dermatologist will consider a list of possible conditions (differential diagnoses) and rule them out one by one, based on the evidence and test results. \\
\\
Instructions: \\
step 1: Evaluate the medical images based on Visual Inspection Guidelines. \\
step 2: Evaluate the medical query as Patient History Guidelines. \\
step 3: Create a case summary using information extracted at step 1 and step 2. \\
step 4: For each candidate skin condition present in the list of Candidates, give a score on a scale of 1 - 10 (where 1 is the least probable and 10 is the most probable) that describes how likely is the given skin condition as a diagnosis for the case summary. \\
step 5: Return the two most probable skin candidates based on scores obtained at step 4. \\
\hline
\end{tabular}
\caption{Expert Guidelines Grounded CoT with Context for Re-Ranker}
\label{tab:with_context_ranking}
\end{table*}

\begin{table*}[ht]
\centering
\begin{tabular}{|p{0.9\textwidth}|}
\hline
You are provided with a dermatology case. For this case, you are provided with some images and list of possible candidates.  \\
\\
Candidates: \textit{<candidates>} \\
\\
Visual Inspection Guidelines: The initial step involves a thorough visual examination of the affected area. The dermatologists keep a track of 10 visual features. \\
\\
1: Size: What is the size of the skin lesions? Is it small or large? \\
2: Shape: What is the shape of the lesions? \\
3: Color: What is the color of the skin lesions? \\
4: Location: Where is the skin lesion or rash located? \\
5: Distribution Pattern: What is the distribution pattern, is it localized or widespread? \\
6: Existence of symmetry: Are the lesions symmetric? \\
7: Borders: Do the edges of the lesion appear sharp, irregular, or blurred? \\
8: Elevation: Is the lesion is flat, raised, or depressed below the skin surface? \\
9: Texture: Does the surface quality looks smooth/scaly/rough/soft/hard. \\
10: Pattern Recognition: Dermatologists are trained in recognizing patterns that certain skin conditions commonly present. These patterns, combined with the other collected information, help in forming a preliminary diagnosis. \\
\\
Act as a dermatologist and follow the instructions below: \\ 
\\
Instructions: \\
step 1: For the given images, use the guidelines and generate a visual description. \\
step 2: For each candidate in the "Candidates", generate the visual description that describes the candidate disease. Also mention distinguishing features based on visual guidelines. Include features like shape, colours, lesion type and area of localization to create a visual description for the disease. \\
step 3: Compare the visual description which was generated for each candidate skin condition at step b with the image description generated at step a. Give a score in the range of 1 to 10 with 1 being the lowest match and 10 being the highest match. \\
step 4: Choose the most probable candidate which has the highest score with the images based on step 3. \\
\hline
\end{tabular}
\caption{Expert Guidelines Grounded CoT without Context for Re-Ranker}
\label{tab:without_context_ranking}
\end{table*}

\begin{table*}[ht]
\centering

\begin{tabular}{|p{0.9\textwidth}|}
\hline
\textbf{Rules:}\\
1. Simplify and Be Direct \\
- Example: “The condition is Chronic Eczema.” \\
- Explanation: Human expert responses tend to be direct and use simpler language. Avoid overly complex explanations and aim for straightforward answers directly addressing the patient’s inquiry. \\
\\
2. Diagnosis Confirmation \\
- Example: “Your diagnosis is a Myxoid Cyst based on the clear image provided.” \\
- Explanation: Include statements that confirm the diagnosis confidently, as seen in responses like “Chronic Eczema.” or “It is myxoid cyst.” Use assertive language to convey confidence in your diagnosis. \\
\\
3. Detail Symptom Correlation \\
- Example: “The semi-spherical cyst near the end of your thumb, as described, leads to a diagnosis of Myxoid Cyst.” \\
- Explanation: Explicitly connect the diagnosis with observed symptoms or test results when applicable, similar to the detailed descriptions in some valid responses. This helps patients understand why a particular diagnosis is made. \\
\\
4. Incorporate Treatment Options Clearly \\
- Example: “For Psoriasis, I recommend oral capsules such as glycyrrhizic acid glycosides, along with transfer factors.”
- Explanation: When suggesting treatments, mention specific medications or procedures clearly and concisely, as observed in responses with high completeness. If possible, explain the purpose of each treatment briefly. \\
\\
5. Mention Commonality or Prevalence \\
- Example: “Chronic Eczema is quite common and effectively manageable with the right treatment.” \\
- Explanation: If applicable, include a brief note on how common the condition is or any relevant statistical information that could reassure the patient or provide context, akin to how some expert responses include prevalence information. \\
\\
6. Use Patient-Friendly Language \\
- Example: “Based on the photo you provided, it looks like you have a Myxoid Cyst, which is a fluid-filled lump that’s not harmful.” \\
- Explanation: Ensure the language used is patient-friendly, avoiding unnecessary medical jargon that could confuse the patient. When medical terms are unavoidable, consider providing a brief, simple explanation. \\
\\
7. Personalization and Empathy \\
- Example: “I understand that dealing with Chronic Eczema can be frustrating. Regular moisturizing and the treatments we’ve discussed should offer relief.”
- Explanation: Whenever possible, personalize the response to the patient’s situation. Display empathy to make your responses feel more human and less robotic. \\

\hline
\end{tabular}
\caption{Automatic Prompt Optimization (APO) Rules}
\label{tab:apo}
\end{table*}

\begin{table*}[ht]
\centering

\begin{tabular}{|p{0.9\textwidth}|}
\hline
\textbf{Rules:}\\
\\
1: Skin condition A is similar to B if they have same name. \\
\\
2: Skin condition A is similar to B if B is also known by the name A. \\
\\
3: Skin condition A is similar to B if both are part of the same root skin condition.\\
Example: Herpetic Eczema and seborrheic eczema are similar since they have same root, Eczema. \\
\\
4: Skin condition A is similar to B if they are both have the same effect and share a common cause. \\
\hline
\end{tabular}
\caption{Evaluation Guidelines Rules}
\label{tab:evaluation_rules}
\end{table*}

\end{document}